\documentclass{article}

\usepackage[final]{neurips_2025}

\usepackage[svgnames]{xcolor} 
\usepackage[utf8]{inputenc} 
\usepackage[T1]{fontenc}    
\usepackage{hyperref}       
\usepackage{url}            
\usepackage{booktabs}       
\usepackage{siunitx}        
\usepackage{amsfonts}       
\usepackage{nicefrac}       
\usepackage{microtype}      
\usepackage{xcolor}         
\usepackage{colortbl}
\usepackage{codebox}
\usepackage{hyperref}
\usepackage{times}
\usepackage{soul}
\usepackage{url}
\usepackage[utf8]{inputenc}
\usepackage[small]{caption}
\usepackage{graphicx}
\usepackage{amsmath}
\usepackage{amsthm}
\usepackage{booktabs}

\usepackage{algpseudocode}
\usepackage{pifont}
\usepackage{amssymb}
\usepackage{mathtools}
\usepackage{balance} 
\usepackage{color}
\usepackage{frame}
\usepackage{tcolorbox}
\usepackage{bm}
\usepackage{inconsolata}
\usepackage{multirow}
\usepackage{subfigure}
\usepackage{float}
\usepackage{wrapfig}
\usepackage{mdframed}
\usepackage{makecell}

\usepackage[ruled,vlined]{algorithm2e}
\usepackage{enumitem}

\usepackage{tcolorbox}
\usepackage{hyperref}
\definecolor{darkmagenta}{rgb}{0.56, 0.0, 1.0}  

\hypersetup{colorlinks,linkcolor={blue},citecolor={darkmagenta},urlcolor={blue}}

\newtheorem{theorem}{Theorem}

\theoremstyle{definition} 

\title{Measure gradients, not activations! Enhancing neuronal activity in deep reinforcement learning}

\author{
  Jiashun Liu\textsuperscript{1}\thanks{Equal contribution. Correspondence to: \texttt{\{ljshasdream, jobando0730\}@gmail.com}} \quad
  Zihao Wu\textsuperscript{1}\footnotemark[1] \quad
  Johan Obando-Ceron\textsuperscript{2,3}\footnotemark[1] \quad
  Pablo Samuel Castro\textsuperscript{2,3} \quad \\
  \textbf{Aaron Courville\textsuperscript{2,3} \quad
  Ling Pan\textsuperscript{1}} \\\\
  \textsuperscript{1} Hong Kong University of Science and Technology \quad \\
  \textsuperscript{2} Mila - Qu\'ebec AI Institute \quad
  \textsuperscript{3} Universit\'e de Montr\'eal
}

\begin{document}

\maketitle

\begin{abstract}

Deep reinforcement learning (RL) agents frequently suffer from neuronal activity loss, which impairs their ability to adapt to new data and learn continually. A common method to quantify and address this issue is the $\tau$-dormant neuron ratio, which uses activation statistics to measure the expressive ability of neurons. While effective for simple MLP-based agents, this approach loses statistical power in more complex architectures. To address this, we argue that in advanced RL agents, maintaining a neuron's \emph{learning capacity}, its ability to adapt via gradient updates, is more critical than preserving its expressive ability. Based on this insight, we shift the statistical objective from activations to gradients, and introduce \texttt{GraMa} (\underline{Gra}dient \underline{Ma}gnitude Neural Activity Metric), a lightweight, architecture-agnostic metric for quantifying neuron-level learning capacity. We show that \texttt{GraMa} effectively reveals persistent neuron inactivity across diverse architectures, including residual networks, diffusion models, and agents with varied activation functions. Moreover, \textbf{re}setting neurons guided by \texttt{GraMa} (\texttt{ReGraMa}) consistently improves learning performance across multiple deep RL algorithms and benchmarks, such as MuJoCo and the DeepMind Control Suite. \textbf{We make our code available\footnote{Code: \url{https://github.com/torressliu/grad-based-plasticity-metrics}}}.
\end{abstract}
\section{Introduction}
\label{sec:introduction}
Deep reinforcement learning (Deep RL) has achieved remarkable success across a variety of domains, including robotics \citep{Liu2021DeepRL}, foundation model fine-tuning \citep{Shao2024DeepSeekMathPT,Liu2025PartIT, Yu2025DAPOAO,Li2025AttentionIL,Lu2025VLARLTM,Liu2025AsymmetricPP}, and game playing \citep{Berner2019Dota2W,pmlr-v202-schwarzer23a}. These advancements have been driven by the expressive power and adaptive learning ability of neural networks which effectively approximate and optimize value functions and/or policies \citep{sokar2023dormant}. However, recent studies have uncovered a critical and often underexplored challenge: as training progresses, subsets of neurons in these networks often experience a progressive loss of activity and become dormant~\citep{sokar2023dormant,ma2024revisiting,qin2024the}. This phenomenon reduces the learning capacity of the network, comprising its ability to adapt to non-stationary data distributions \citep{Nikishin2022ThePB}, which in turn hinders their ability to acquire new knowledge and adapt to evolving environments \citep{Abbas2023LossOP}. Despite its importance, quantifying and mitigating neuronal activity remains challenging due to its complex underlying mechanisms \citep{understand,obando2023small,overfit, Lyle2024DisentanglingTC,liu2025the}.

To address this problem, a primary principle has been to restore a network's learning ability by reactivating or resetting inactive neurons. These approaches span multiple granularities: model-level and layer-level resets that reinitialize specific layers, while straightforward, often lead to catastrophic forgetting~\citep{nikishin2023deep}. On the other hand, neuron-dependent resets (e.g., ReDo~\citep{sokar2023dormant}) target specific underperforming neurons \citep{Xu2023DrMMV} and provide a finer-grained approach by selectively reinitializing a subset of neurons identified as dormant, which mitigates forgetting and maintains computational efficiency.

The effectiveness of neuron-dependent resets hinges critically on having a reliable criterion to identify which neurons require initialization. Existing methods primarily rely on activation-based metrics, such as the $\tau$ dormant neuron metric~\citep{sokar2023dormant}, which measures neuronal inactivity based on activation values (i.e., the output of activation functions). 
It has demonstrated utility in standard architectures, and has been used to guide targeted neuron resets to restore activity in simple settings such as serial MLPs with ReLU activations~\citep{Agarap2018DeepLU}, providing a simple means of maintaining learning capability without relying on auxiliary networks \citep{nikishin2023deep} or models \citep{Lee2024SlowAS}. 
By identifying neurons with weak or no activation and selectively reinitializing them, the activation-based dormancy metrics become a widely adopted tool for restoring learning capabilities and preventing performance plateau in standard deep RL settings~\citep{farias2025selfnormalized, liu2025neuroplastic, juliani2024a}.

\begin{wrapfigure}{r}{0.5\textwidth}
\centering
\vspace{-0.7cm}
\includegraphics[width=\linewidth]{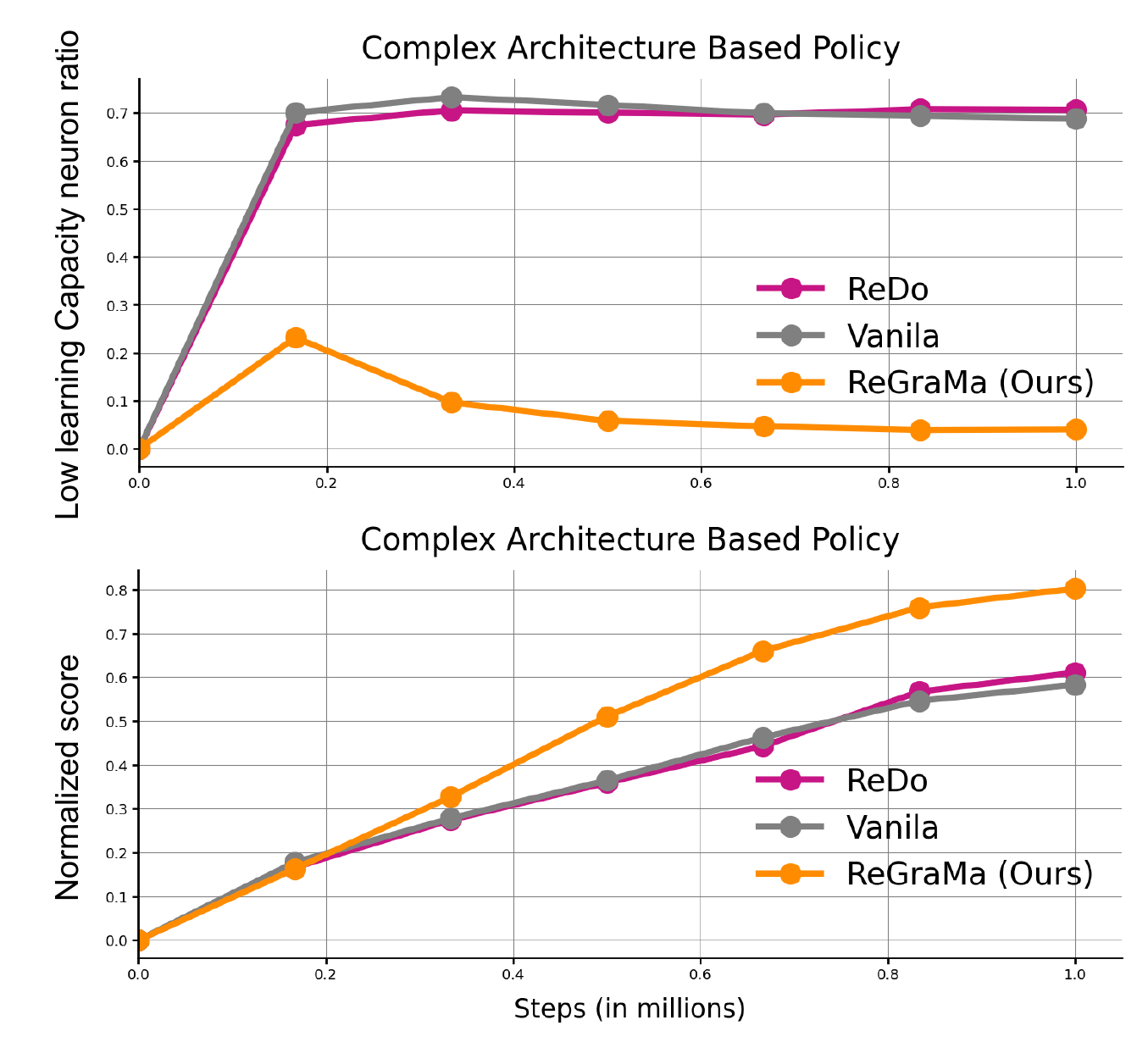}
\vspace{-0.6cm}
\caption{\textbf{Dormant neuron metric struggles in advanced vision RL BRO-net agent~\citep{nauman2024bigger}}. Neuron resetting based on the dormant neuron index (ReDo) cannot restore the learning capacity of the agent, which limits its effectiveness. The curves record normalized score across 3 image-input tasks (15 runs per method), i.e., Dog Stand, Dog Walk, Dog Run.}
  \label{fig_1}
\vspace{-0.4cm}
\end{wrapfigure}

However, deep RL architectures have rapidly evolved beyond these simple architectures, integrating advanced components such as residual connections~\citep{nauman2024bigger,lee2025simba}, mixture of experts (MoEs)~\citep{obando2024mixtures,sokar2024don,willimixture}, and diffusion-based models \citep{ren2025diffusion, wang2024diffusion} to improve scalability and performance for tackling complex tasks involving continuous control and high-dimensional observations. Additionally, they also imperceptibly alter the statistical properties of neuron activations, creating a significant mismatch with traditional activity metrics. 
Through our extensive analysis, we find that resetting based on $\tau$-dormant neuron ratio (ReDo) reduces effectiveness in these newer architectures. 

As shown in \autoref{fig_1}, ReDo struggles to identify inactive neurons that have lost their learning capacity in the more advanced BRO-net architecture, which features residual paths and layer normalization, ultimately failing to achieve the intended enhancement.
This mismatch arises because activation-based metrics focus solely on a neuron's current output, evaluating its expressive capacity (how strongly it activates), while neglecting its learning capacity (how effectively it can adapt to new data distributions). 
Consequently, it becomes particularly problematic as deep RL architectures evolve beyond simple feedforward networks to incorporate advanced structural elements, diverse activation functions, and sophisticated normalization techniques. In these modern architectures, a neuron's activation magnitude often fails to accurately reflect its true learning potential.

To address this mismatch, we propose a fundamental shift in perspective: from evaluating neurons based on their outputs to evaluating their learning potential by leveraging gradient magnitude.
While activation values only capture what a neuron currently expresses, gradients tend to measure a neuron's capacity in response to the given situation that directly drives parameter updates.
This makes it a general and natural proxy for neuronal health across architectural variations. Building upon this insight, we introduce \texttt{GraMa} (\textbf{Gra}dient \textbf{Ma}gnitude based Neuronal Activity Metric), a robust and lightweight framework for quantifying neuronal activity via the gradient magnitudes. \texttt{GraMa} maintains validity across diverse architectural patterns, making it well-suited for modern deep RL agents.

In addition, \texttt{GraMa} imposes negligible computational and memory overhead by utilizing information already present in the optimization pipeline, which is a critical consideration for resource-intensive RL training. Leveraging \texttt{GraMa}'s efficiency, we develop a targeted neuron reset mechanism, (\texttt{ReGraMa}), that selectively reinitializes inactive neurons that have lost their learning capacity during training. This mechanism demonstrates robust efficiency across various architectures, including the SAC variant \citep{nadimpalli2025evolving}, the residual BRO-net \citep{nauman2024bigger}, and the diffusion-based policy DACER \citep{wang2024diffusion}.

\begin{tcolorbox}[colback=blue!4,
leftrule=0.5mm,top=0mm,bottom=0mm] \textbf{Our contributions are summarized as follows:}
\begin{itemize}[leftmargin=*]
    \item We show that the widely-adopted activation-based neuronal health measurements lose statistical power in complex architectures and provide a qualitative analysis of the underlying causes.
    \item We reframe neuronal health evaluation through \texttt{Grama}, a gradient-based metric that quantifies learning potential independently of architectural complexity, and demostrate that neuronal learning capacity degradation affects even state-of-the-art network developments.
    \item We develop (\texttt{ReGraMa}), an efficient neuron resetting mechanism guided by \texttt{GraMa}, which effectively restores neuronal activity across a wide range of network architectures.
    \item We conduct extensive experiments on MuJoCo \citep{Brockman2016OpenAIG}, DeepMind Control Suite \citep{Tassa2018DeepMindCS}, showing that \texttt{GraMa}-guided resetting improves performance and learning stability across diverse architectures.
\end{itemize}
\end{tcolorbox}
\section{Background}\label{sec.back}
A reinforcement learning (RL) problem is typically formalized as a Markov Decision Process (MDP), defined by the tuple $(\mathcal{S}, \mathcal{A}, \mathcal{P}, \mathcal{R}, \gamma)$, where $\mathcal{S}$ is the state space, $\mathcal{A}$ the action space, $\mathcal{P}$ the transition probability function $\mathcal{S} \times \mathcal{A} \times \mathcal{S} \to [0, 1]$, $\mathcal{R}$ the reward function $\mathcal{S} \times \mathcal{A} \to \mathbb{R}$, and $\gamma \in [0, 1)$ the discount factor. The state-action value function under policy $\pi$ is given by:$Q^{\pi}(s, a) = \mathbb{E}_{\pi} \left[ \sum_{t=0}^{\infty} \gamma^t \mathcal{R}(s_t, a_t) \mid s_0 = s, a_0 = a \right].$ The objective is to find a policy $\pi$ that maximizes the expected return $Q^{\pi}(s, a)$ for each state $s$. In actor-critic-based deep RL, the Q-function is approximated by a neural network \( Q_\theta \) with parameters \( \theta \). During training, the agent interacts with the environment and stores trajectories in a replay buffer \( D \). Mini-batches sampled from \( D \) are used to update \( Q_\theta \) by minimizing the temporal difference loss:$\mathcal{L}_{Q}(\theta) = \mathbb{E}_{(s, a, r, s') \sim D} \left[(Q_{\theta}(s, a) - Q^{\mathcal{T}}(s, a))^2\right],$ where the target is given by \( Q^{\mathcal{T}}(s, a) = \mathcal{R}(s, a) + \gamma Q_{\bar{\theta}}(s', \pi_{\bar{\phi}}(s')) \), and \( \bar{\theta}, \bar{\phi} \) denote the parameters of target critic and actor networks, respectively.

\paragraph{Neuronal activity measurement based on activation value.}
Recent studies have identified that the dynamic and non-stationary nature of RL objectives can cause neurons to permanently lose their activity~\citep {Lu2018CollapseOD}, impairing the network's ability to fit new data and thereby limiting learning progress~\citep{Nikishin2022ThePB,ceron2024value}. This phenomenon, referred to as the \emph{dormant neuron phenomenon}, has been quantitatively characterized using the \emph{$\tau-$dormant neuron ratio}~\citep{sokar2023dormant}, and serves as a core metric for assessing neuron-level plasticity~\citep{Xu2023DrMMV,qin2024the,liu2025neuroplastic}. Specifically, a neuron $i$ in layer $\ell$ is considered $\tau$-dormant if its normalized activation (see \autoref{eq.3}) falls below a threshold $\tau$, where $H^\ell$ is the number of neurons in layer $\ell$, and $D$ denotes the data distribution. $h_i^\ell(x)$ represents the activation value of neuron $i$ given input $x$.
\begin{equation}\label{eq.3}
S_i^\ell = \frac{\mathbb{E}_{x \sim D} \left| h_i^\ell(x) \right|}{\frac{1}{H^\ell} \sum_{k \in h} \mathbb{E}_{x \sim D} \left| h_k^\ell(x) \right| }
\end{equation}\section{Related Work}\label{related}
Dynamic objectives may cause irreversible damage to the neuronal activity of networks during training \citep{understand,overfit,castanyer2025stablegradientsstablelearning}, which is also significant in the field of multi-agent RL \citep{qin2024the} and visual deep RL \citep{ma2024revisiting}. This issue may lead to the phenomenon where deep learning agents progressively lose their ability to fit new data~\citep{Abbas2023LossOP}, and limit their capacity for continual learning \citep{elsayed2024addressing}. In the field of Deep RL, there are some factors that have been found to have a direct correlation with the activity of neurons \citep{juliani2024a,mayor2025the,castanyer2025stablegradientsstablelearning}. Among these, activation functions play a particularly important role, recent findings indicate that replacing ReLU activation can help preserve neuron-level learning capacity~\citep{Abbas2023LossOP}.

Certain loss functions, especially those incorporating L2 regularization, have been found to mitigate the network's activity degradation~\citep{Kumar2023MaintainingPI}. Another relevant line of research explores architectural adjustments such as scaling up~\citep{nikishin2023deep} or topology growth~\citep{liu2025neuroplastic}. These methods can introduce modules that enhance the proportion of active, high-quality neurons. A separate and increasingly influential line of research focuses on directly manipulating model parameters to recover their activity. Resetting neural networks, either periodically or selectively, has been shown to be both effective and easy to implement~\citep{farias2025selfnormalized}. For instance, \citet{Nikishin2022ThePB} demonstrates the benefits of periodic resets for continual learning. 

Similarly, \citet{ma2024revisiting} shows that selectively resetting specific layers can improve stability without sacrificing expressive capacity. At a finer granularity, ReDO~\citep{sokar2023dormant} introduces a neuron-level resetting scheme that outperforms earlier coarse strategies in terms of stability and precision. However, ReDO’s reliance on activation-based quantization limits its applicability to more complex architectures~\citep{understand,Lyle2024DisentanglingTC}. To address this limitation, we introduce a novel neuronal activity quantification approach based on \textit{gradient magnitudes}, enabling generalized activity estimation and recovery across arbitrary network designs. Notably, \citet{Ji2024ACEO} provided the first empirical evidence in model-free reinforcement learning of a strong correlation between the internal gradient dynamics of the agent’s policy network and its learning capacity, offering compelling motivation for the present study.
\section{\underline{Gra}dient \underline{Ma}gnitude based Neuronal Activity Metric (\texttt{GraMa})}\label{sec:gate}

In this section, we qualitatively investigate ReDo's limitations from the perspective of architectural composition \autoref{sec.3.1}. We then introduce \texttt{GraMa} (\autoref{sec.3.2}), a novel neuronal activity metric that redefines the statistical objective from activation values to gradient magnitudes. Next, we perform an in-depth analysis of the characteristics of low learning capacity neurons using \texttt{GraMa}, and provide both theoretical and empirical analyses of the performance similarity between \texttt{GraMa} and ReDo on simple architectures. Finally, we analyze the advantages of \texttt{GraMa} in complex architectures.
\begin{figure}[t] 
\centering
\vspace{-0.5cm}
\includegraphics[width=1\textwidth]{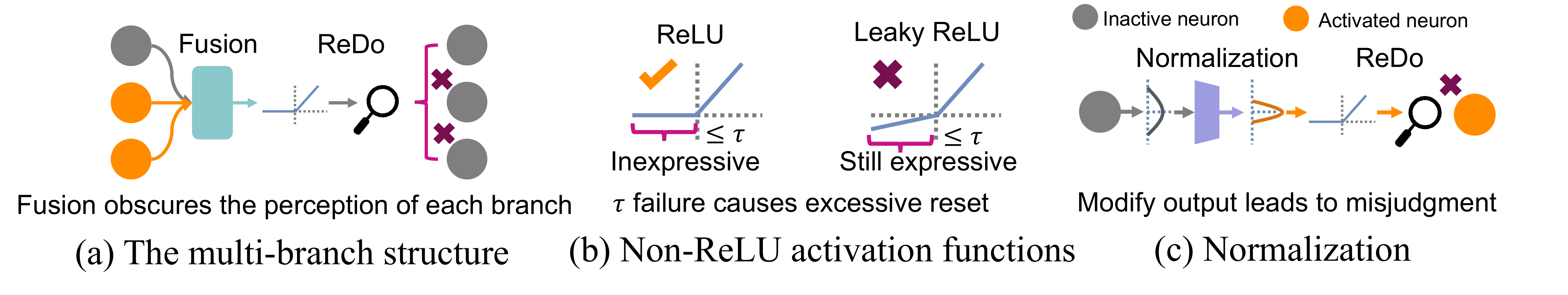}
\caption{\textbf{Key techniques involved in advanced models may reduce the efficiency of activation-based metrics.} (a) The final activation values observed after branch fusion no longer accurately reflect the individual contributions of neurons in each branch. (b) A near-zero value fails to represent the expressivity with non-ReLU activation function. (c) Normalization confuses ReDo by modifying the neuron's outlier output.}
\label{fig.case}
\vspace{-0.15cm}
\end{figure}

\subsection{Misalignment of activation-based neuron activity detection in modern architectures}\label{sec.3.1}
\begin{wrapfigure}{r}{0.54\textwidth}
\centering
\vspace{-0.6cm}
\includegraphics[width=\linewidth]{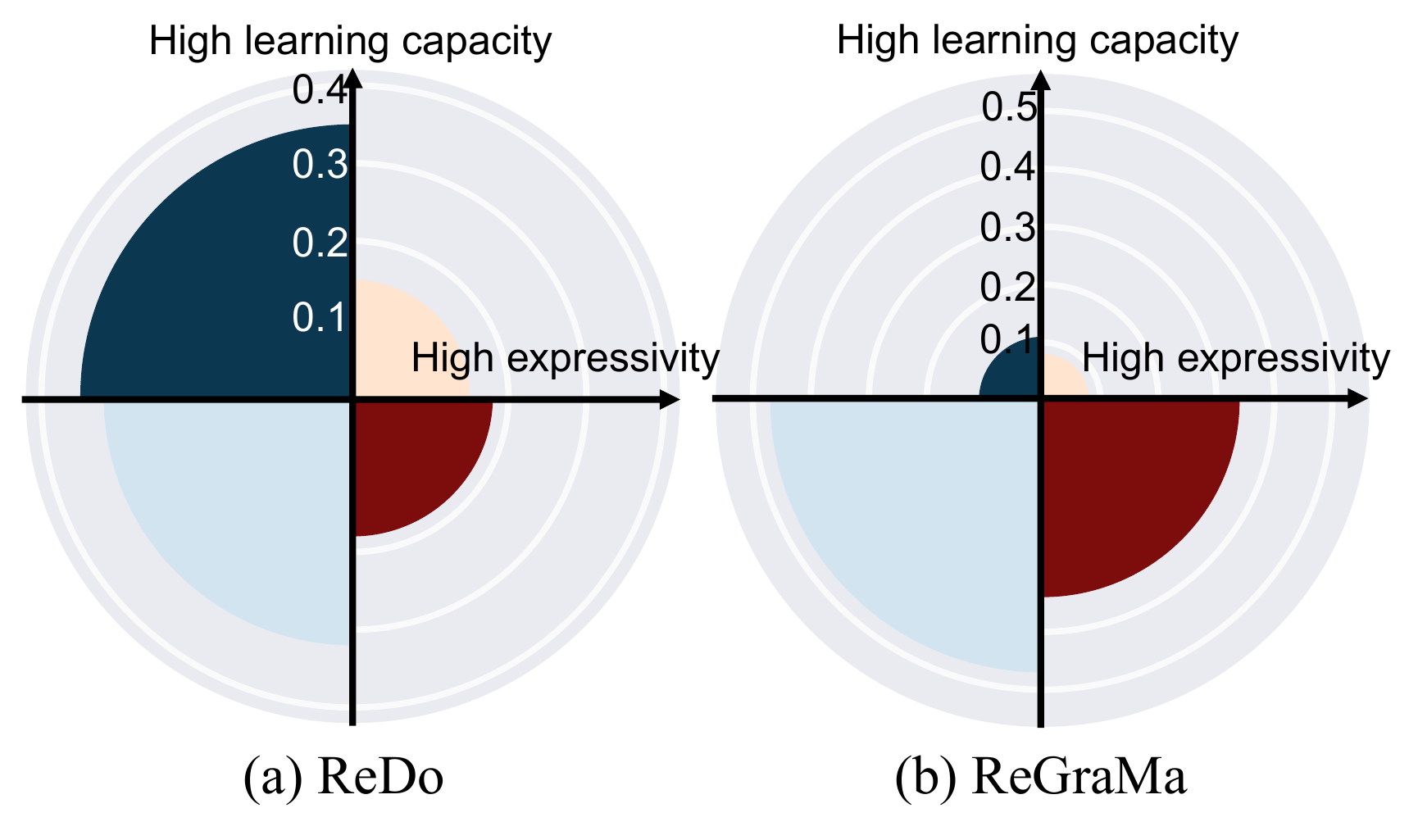}
\vspace{-0.6cm}
\caption{\textbf{ReDo fails to reset neurons with low learning capacity (\textcolor{red}{red}), while also incorrectly resets the neurons with high learning capacity (\textcolor{blue}{dark blue})}, including those with both high learning capacity and high expressiveness (white). This plot shows the proportion of resetting 4 neuron types during the same reset step in the Dog Walk task.}
  \label{fig.wrong_id}
\vspace{-0.5cm}
\end{wrapfigure}The widely adopted activation-based neuron activity measures~\citep{sokar2023dormant} operate on the assumption that a neuron's activation magnitude directly corresponds to its contribution to learning.
Motivated by the development of more complex network structures in language and vision domains, recent trends in deep RL have moved beyond simple serial MLP architectures with ReLU activations towards more sophisticated network designs for scaling. 
However, this assumption becomes increasingly fragile in modern complex architectures which are rapidly becoming standard practice. 
We reveal a misalignment between activation values and actual learning potential, leading to inefficient neuron identification.

\paragraph{Motivating example.} To systematically analyze this misalignment, we evaluate BRO-net~\citep{nauman2024bigger}, a recently proposed architecture, on the challenging Dog Walk task \citep{Tassa2018DeepMindCS}. 
We categorized neurons into four quadrants based on their expressive capacity (measured by activation magnitude) and learning capacity (measured by gradient magnitude), which we introduce in more detail in \autoref{sec.3.2}.
Neurons are ranked according to these two criteria, and we select the top 25\% of neurons in each ranking, so as to provide an understanding of the types of neurons reset by each method.
The results in \autoref{fig.wrong_id} demonstrate that ReDo cannot accurately identify neurons that have genuinely lost their learning capacity (\textcolor{red}{red}) based on their activation values. More importantly, ReDo mistakenly resets a significant number of neurons with high learning capacity that are less expressive at the moment (\textcolor{blue}{dark blue}).

\begin{figure}[ht] 
\centering
\hspace{-0.5cm}\includegraphics[width=1.0\linewidth]{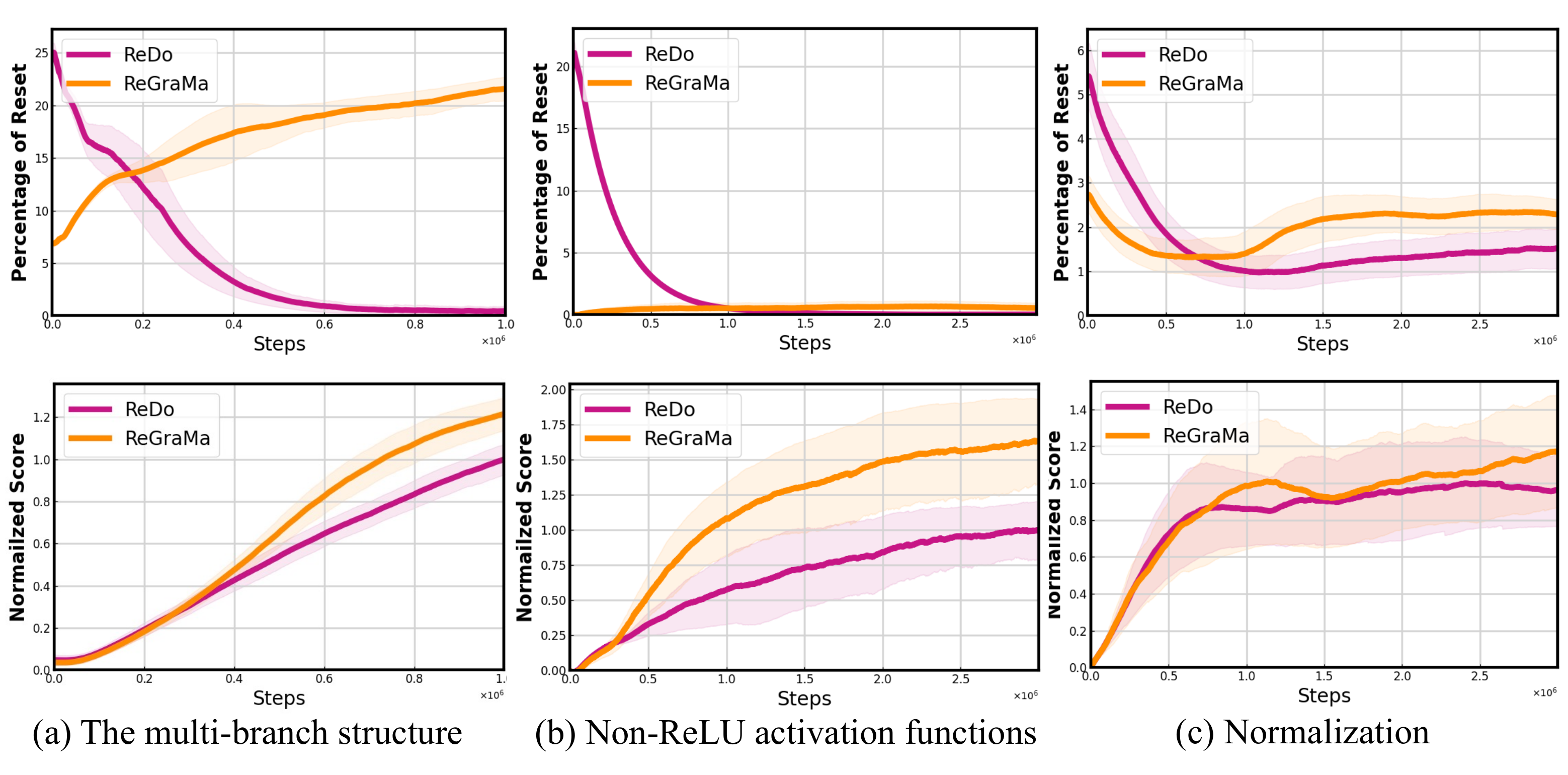}%
\caption{\textbf{Empirical validation corresponding to the three cases of \autoref{fig.case}}. Top row records the proportion of reset neurons. Bottom row shows the performance. Each curve represents the average over 3 seeds (Dog Walk).}
\label{fig.caseresults}
\vspace{-0.5cm}
\end{figure}

\paragraph{Analysis.} In addition to the primary factor of statistical objectives, we identify three key architectural features that may undermine the reliability of activation-based neuron dormancy detection. \autoref{fig.case} provides an intuitive illustration of each. (i) \textit{Multi-branch network structures.} In modern architectures like ResNets~\citep{He2015DeepRL}, information flows through multiple branches before being fused and passing through activation functions. This architectural pattern introduces a critical problem: the final activation values observed after branch fusion no longer accurately reflect the individual contributions of neurons in each branch.
Our experiments with Resnet-SAC \citep{Shah2021RRLRA} shown in \autoref{fig.caseresults} (a) confirm this effect, where the results suggest that activation-based methods lose their ability to identify dormant neurons and limits the performance of the agent.
(ii) \textit{Non-ReLU activation function.} The interpretability of activation values is highly dependent on the specific activation function used. While ReLU creates a clear distinction between ``dead'' (output=$0$) and ``active'' neurons, this clarity breaks down when considering alternative activation functions. For example, with Leaky ReLU, neurons may remain expressive and contribute to learning even when their pre-activation values are negative. 

As a result, solely relying on activation values to measure neuron dormancy becomes unreliable and may lead to misidentification. Our experiments replacing ReLU with Leaky ReLU in SAC (\autoref{fig.caseresults} (b)) demonstrate this problem, where the activation-based method (ReDo) shows considerable reset activity during early learning with lower performance, indicating that it struggles to establish meaningful dormancy criteria. 
(iii) \textit{Normalization layers.} Normalization techniques modify the distribution of neuron outputs before they pass through activation functions. This process adjusts outliers and rescales values across entire layers, causing post-normalization values to lose a direct correspondence with individual neuron functionality. Thus, activation values measured after normalization may no longer accurately reflect a neuron's learning capacity. Our experiments incorporating layer normalization into SAC, \autoref{fig.caseresults} (c) reveal that the activation-based method struggles to maintain consistent neuron assessment, whereas our method consistently identifies relevant neurons throughout the learning process.
\subsection{\underline{Gra}dient \underline{Ma}gnitude based Neuronal Activity Metric (\texttt{GraMa})}\label{sec.3.2}
Learning activity is directly related to the gradient received by a neuron during training \citep{zhou2024how}. A natural and simple way to enhance ReDo-like metrics is to count how many neurons in each network layer fail to obtain meaningful gradients from the current sample batch.
Given an input distribution $D$, let \colorbox{AntiqueWhite}{$|\nabla_{h_i^l}L(x)|$} denote the gradient magnitude of the neuron $i$ in layer $\ell$ under input $x \in D$ and $H^\ell$ be the number of neurons in layer $\ell$. Based on \autoref{eq.3}, we compute a learning capacity score for each neuron using the normalized average of the corresponding layer $\ell$, as shown in \autoref{eq.4}.
\begin{equation}\label{eq.4}
G_i^\ell=\frac{\mathbb{E}_{x\in D}\colorbox{AntiqueWhite}{$|\nabla_{h_i^l}L(x)|$}}{\frac{1}{H^l}\sum_{k\in h}\mathbb{E}_{x\in D}\colorbox{AntiqueWhite}{$|\nabla_{h_i^l}L(x)|$}}.
\end{equation}

\begin{wrapfigure}[13]{R}{0.52\textwidth}
\begin{minipage}{0.52\textwidth}
\vspace{-0.5cm}
\begin{algorithm}[H]
	\caption{\texttt{ReGraMa}}
	\label{alg.1}
        \SetKwInOut{Input}{Input}
        \Input{Model $\theta$, threshold $\tau$, frequency $\Delta_t$} 
        \While{$t<$ maximum training time}{ 
            Update $\theta$ with regular RL loss\;
            \If{$t\mod \Delta_t==0$}{
                \For{each layer $\ell$}{
                    \For{eachneuron $i$}{
                        Calculate $G_i^\ell$\Comment{\autoref{eq.4}}\\
                        \If{$G_i^\ell\leq\tau$}{
                            Reinitialize neuron $i$\;
                        }
                    }
                }
            }
        }
\end{algorithm}
\end{minipage}
\end{wrapfigure}

\begin{figure}[t] 
\centering
\vspace{-0.5cm}
\includegraphics[width=0.32\linewidth]{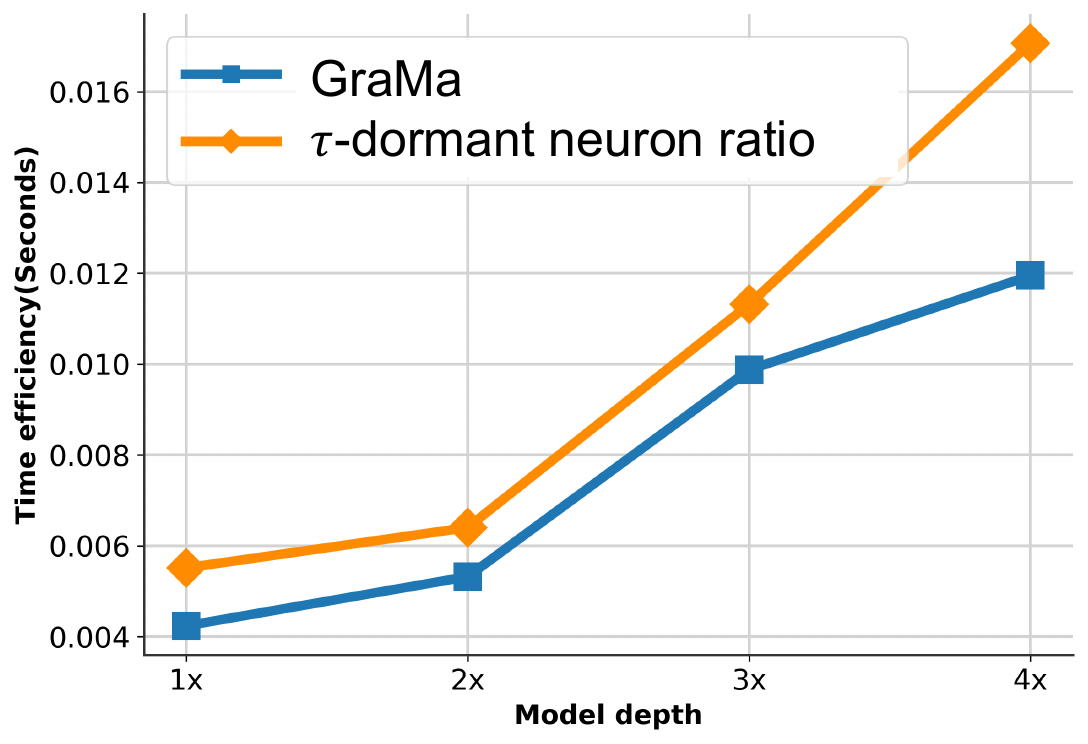}%
\vline height 92pt depth 0 pt width 1.5 pt
\includegraphics[width=0.65\linewidth]{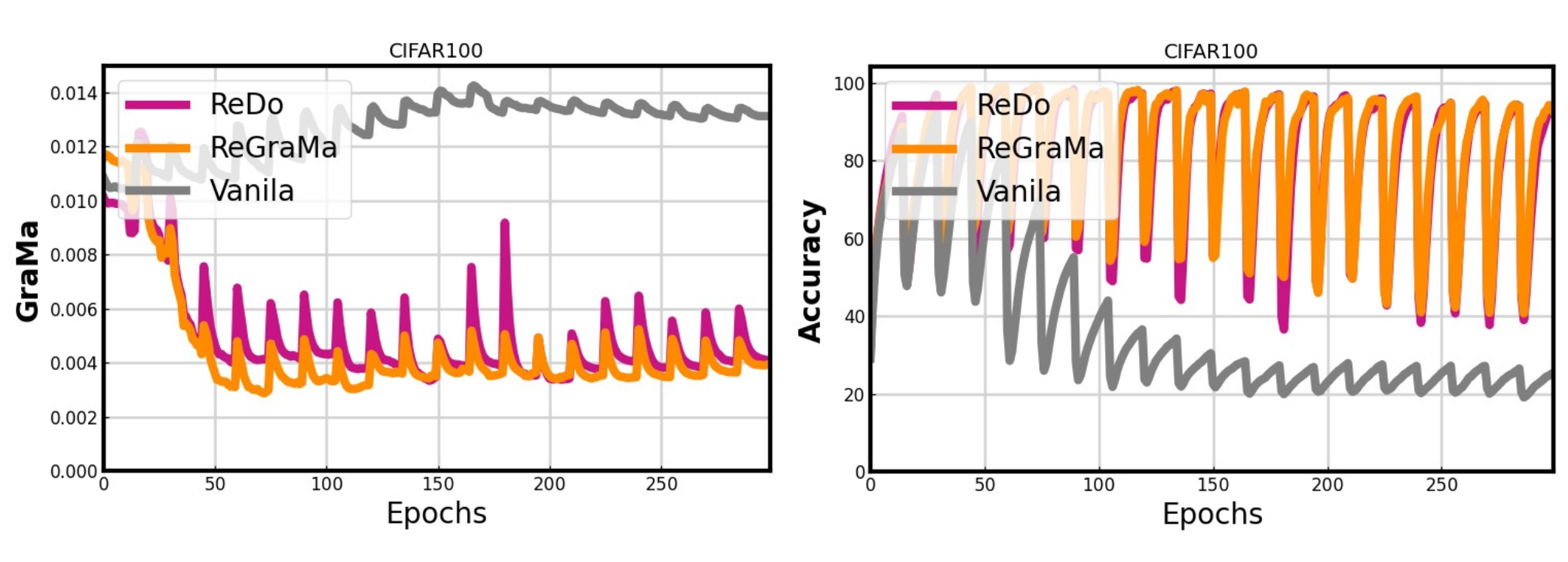}
\caption{(Left) Execution time comparison based on BRO-net (RTX3090 GPU); (Right) The results verify that \texttt{ReGraMa} is as effective as ReDo on traditional network architectures. The backbone, hyperparameter settings and number of seeds are the same for all experiments.}
\label{fig_time}
\end{figure}

\textbf{\texttt{GraMa} determines that neuron $i$ in layer $\ell$ is inactive when $G_i^\ell \leq \tau$.}
\texttt{GraMa} has the same form as the dormant neuron ratio, but redefines its core signal, using gradient magnitudes $|\nabla_{h_i^l}L(x)|$ rather than activation values $h_i^l(x)$. 
The pre-set threshold $\tau$ allows us to detect neurons with outlier gradient magnitude. 
Since gradient information is available in the tensor after backpropagation at each step, there is no need to store the intermediate outputs of each neuron during forward computation, which is required by activation-based metrics. This makes \texttt{GraMa} lightweight, as verified in \autoref{fig_time} (left).
\paragraph{Resetting neurons guided by \texttt{GraMa} (\texttt{ReGraMa}).} Neuron resetting is a simple technique widely used to preserve the learning capacity of deep RL agents \citep{Nikishin2022ThePB}. Our approach follows the ReDo pipeline ~\citep{sokar2023dormant}, as outlined in Algorithm~\ref{alg.1}: during training, we periodically quantify the activity of neurons in all layers using \texttt{GraMa}; any neuron $i$ with $G_i^\ell \leq \tau$ is considered inactive and reinitialized. Specifically, reinitialization involves resetting incoming weights to the original weight distribution, while outgoing weights are set to zero. It is worth noting that, empirically, we found that although LayerNorm layers contain only two learnable parameters, they tend to be quite inactive. As a result, \texttt{ReGraMa} is deployed for these parameters in our official implementation, leading to performance improvements that warrant further analysis. Another avenue worth exploring is the potential for resetting the input weights and biases of neurons.

\begin{figure}[ht]
    \centering
    \vspace{-0.3cm}
    \subfigure[]{
        \includegraphics[width=0.45\textwidth]{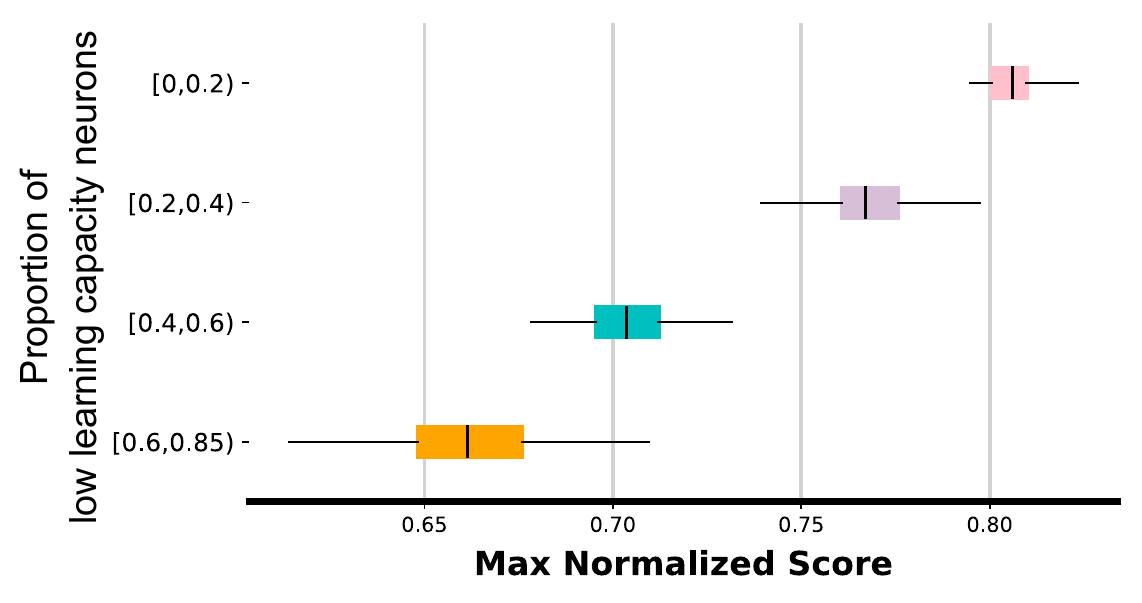}
        \label{fig:figure1}
    }
    \hfill
    \subfigure[]{
        \includegraphics[width=0.45\textwidth]{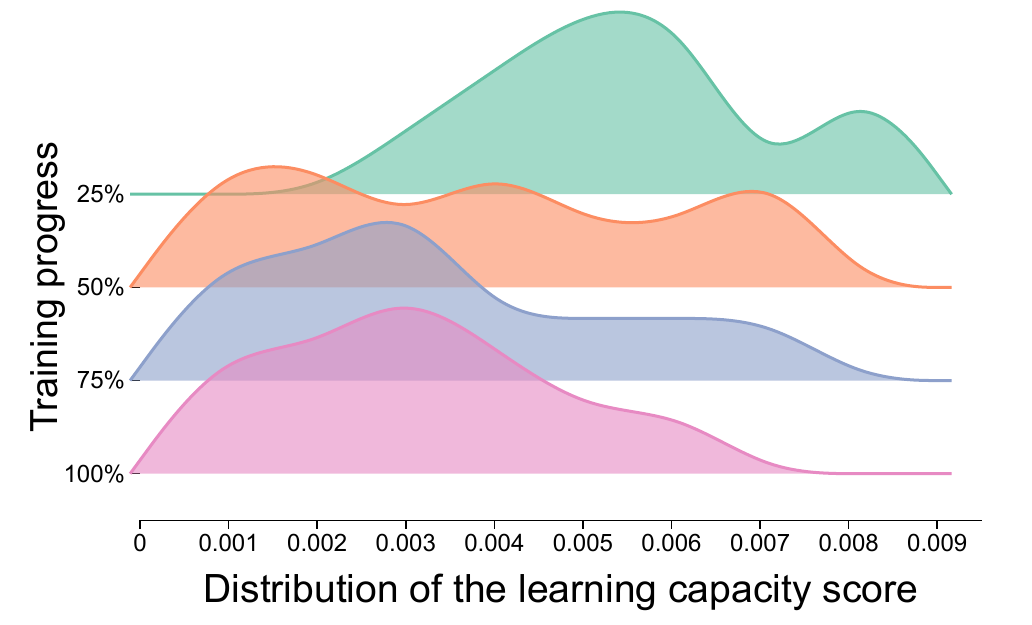}
        \label{fig:figure2}
    }
    \caption{(a): \textbf{Low learning capacity neurons have a direct negative impact on agent performance.} X-axis represents the normalized performance, and the Y-axis denotes the BRO-nets with different proportions of low learning capacity neurons. (b): \textbf{Neuron loss of learning capacity is irreversible.} x-axis denotes the number of neurons under each \texttt{GraMa} score, and the y-axis shows the training phase. (define $>0.0095$ as active). }
    \label{fig:two_figures}
\end{figure}

\paragraph{The ratio of low learning capacity neurons is inversely related to performance.} We control the number of reset neurons in \texttt{ReGraMa} to evaluate the performance of four agents with varying proportions of low learning capacity neurons on the Dog Walker task. Results in \autoref{fig:two_figures} (a) indicate that performance degrades significantly as the ratio of low learning capacity neurons increases.
\paragraph{Neuron loss of learning capacity is irreversible.} We use \texttt{GraMa} with a threshold of $\tau=0.0095$ to sample $1000$ inactive neurons in the pre-training period of vanilla agent. We then trace the change of their learning capacity scores and analyze the score distribution over time, as shown in \autoref{fig:two_figures} (b). The results indicate that none of the sampled neurons exceed the threshold of $0.0095$ as training progresses. This suggests that such neurons are unable to recover their learning ability independently, further underscoring the importance of resetting neurons with low learning capacity.
\paragraph{Equivalence of \texttt{ReGraMa} to ReDo in traditional architectures.} 
We theoretically analyze the similarities between \texttt{ReGraMa} and ReDo in traditional architectures (MLP with ReLU activations~\citep{sokar2023dormant}) and draw the following conclusions:
\begin{theorem}
If neuron $i$ is dormant (\( s_i^{\ell} = 0 \)), then both \( \nabla_{h_i^{\ell}} f = 0 \) and $G_i^\ell = 0$.
\end{theorem}
\textit{Proof. }From the dormant neuron formula (\autoref{eq.3}), we can conclude that:
\[s_i^{\ell} = 0 \quad \iff \quad \mathbb{E}_{x \in D} |h_i^{\ell}(x)| = 0.\] Since \( |h_i^{\ell}(x)| \geq 0 \), \( |h_i^{\ell}(x)|=0 \iff\;h_i^{\ell}(x) = 0\). This means the neuron is almost never activated on the dataset \( D \). The derivative of the ReLU \citep{Agarap2018DeepLU} activation function is:
\[
\frac{\partial h_i^{\ell}(x)}{\partial z_i^{\ell}(x)} = 
\begin{cases}
1 & \text{if } z_i^{\ell}(x) > 0, \\
0 & \text{if } z_i^{\ell}(x) \leq 0.
\end{cases}
\]
$z$ represents the output after passing through the activation function. Thus, if \( h_i^{\ell}(x) = 0 \), then \( z_i^{\ell}(x) \leq 0 \), and during backpropagation, the gradient turns to zero:
\[
\nabla_{z_i^{\ell}} f = \mathbb{E}_{x \in D} \left[ \frac{\partial f}{\partial h_i^{\ell}(x)} \cdot \frac{\partial h_i^{\ell}(x)}{\partial h_i^{\ell}} \right] = \mathbb{E}_{x \in D} \left[ \frac{\partial f}{\partial h_i^{\ell}(x)} \cdot \underbrace{\frac{\partial h_i^{\ell}(x)}{\partial z_i^{\ell}(x)}}_{=0} \cdot h^{\ell-1}(x) \right] = 0.
\]
\begin{tcolorbox}[colframe=orange!60!, colback=white,
leftrule=0.5mm,top=0mm,bottom=0mm]
\textbf{Takeaway. }\textit{We prove that, in the traditional architecture (MLP with ReLU), neurons that are identified as inactivate by ReDo will also be identified as such by \texttt{ReGraMa}. }
\end{tcolorbox}

\paragraph{Empirical verification.} We trained a traditional fully connected network with ReLU on the CIFAR-100 benchmark \citep{Krizhevsky2009LearningML}, following the continuous learning experimental setup in \cite{Dohare2024LossOP}. Every 15 epochs, a new category of data is added to the training set, requiring the agent to classify samples from the whole data distribution. The two \textcolor{gray}{\textbf{gray curves}} in \autoref{fig_time} (right) show that the vanilla agent's accuracy gradually declines as training progresses, indicating a loss of learning ability. 
Meanwhile, the proportion of inactive neurons detected by \texttt{GraMa} increases over time and fluctuates with the same periodicity as the accuracy curve, with both stabilizing around epoch 150. 
To further validate \texttt{ReGraMa}, we conducted a two-part intervention study: (1) We reset neurons identified as inactive by the $\tau-$dormant neuron ratio (ReDo) whenever a new data category is introduced. The resulting \textcolor{magenta}{\textbf{pink curves}} in \autoref{fig_time} show that resetting dormant neurons can both improve performance and reduce \texttt{GraMa} ratio. (2) We then reset neurons flagged by \texttt{ReGraMa} as having low activity. This intervention produced improvements similar to those of ReDo, as illustrated by the \textcolor{orange}{\textbf{orange curves}} in \autoref{fig_time}.  
These results suggest that \texttt{ReGraMa} is as effective as the ReDo metric in identifying neuronal inactivity in simple network architectures.

\paragraph{\texttt{ReGraMa} identifies inactive neurons more effectively in complex architectures.}
\begin{wrapfigure}{r}{0.5\textwidth}
\centering
\vspace{-0.3cm}
\includegraphics[width=\linewidth]{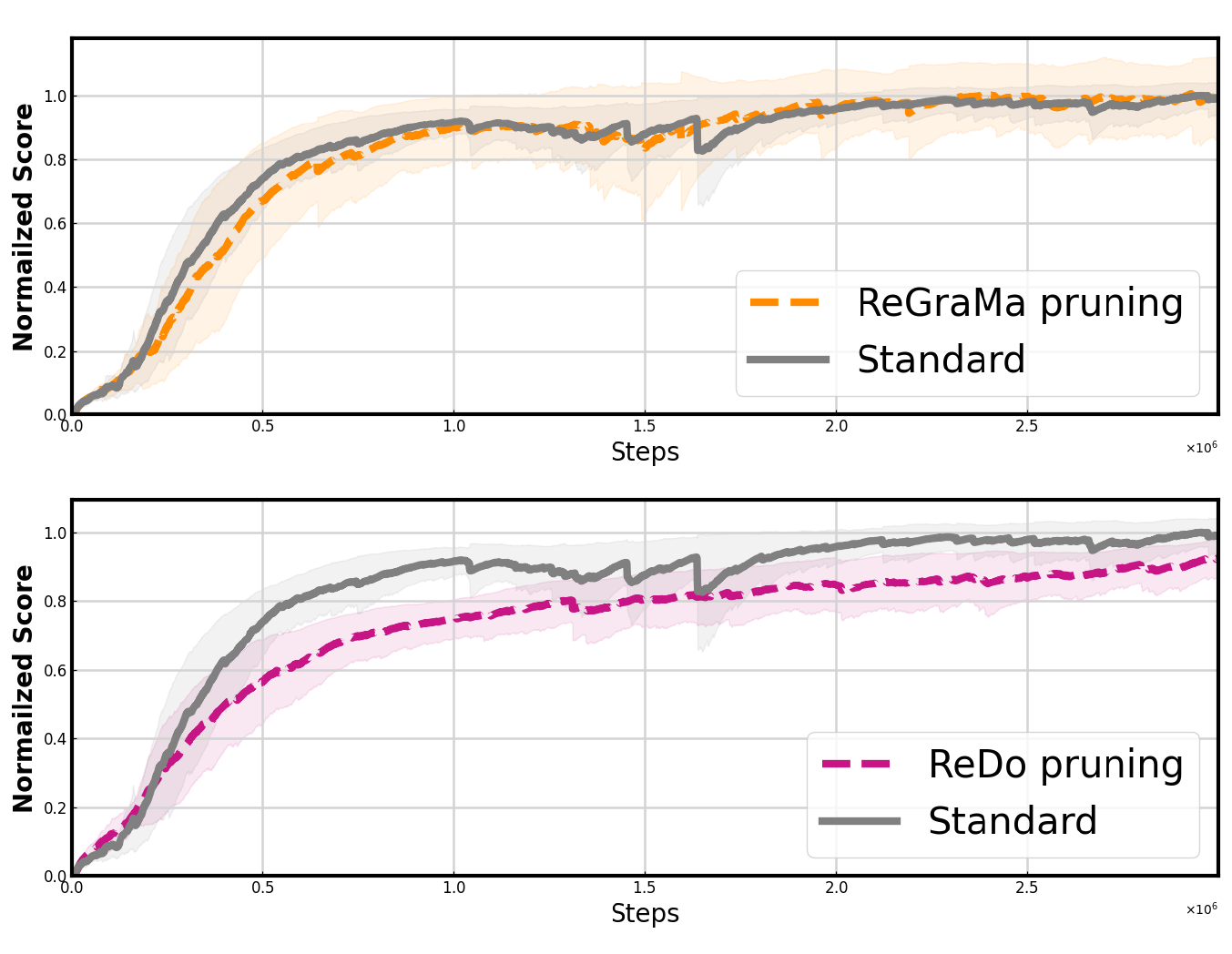}
\caption{\textbf{Pruning neurons identified as inactive by \texttt{ReGraMa} during the training has less impact on the performance in Dog Walk.} Standard denotes vanilla agent. Curves show the average over four seeds.}
\label{fig.prune}
\vspace{-0.2cm}
\end{wrapfigure}When moving to more complex architectures, resetting neurons based solely on activation values becomes less reliable. In deep networks with normalization layers, residual connections, or context-dependent features, activations can fluctuate substantially even for neurons that are not meaningfully contributing to learning. As a result, activation magnitude alone is a noisy indicator of long-term inactivity. To examine this limitation, we compared activation-based and gradient-based criteria by pruning the corresponding neurons from a complex \texttt{BRO-net} agent and evaluating the resulting performance. 

The degree of post-pruning performance degradation reflects both the relevance of the pruned neurons and the reliability of the underlying metric. As shown in \autoref{fig.prune}, pruning neurons identified by \texttt{ReGraMa} causes only minimal performance loss, indicating that gradient magnitude provides a more stable and task-relevant measure of neuronal inactivity. This finding suggests that \texttt{ReGraMa} detects structural inactivity—neurons that remain consistently uninformative during optimization—rather than transient activation noise. Such stability is particularly advantageous in large, modular architectures, where distinguishing genuine inactivity from context-specific silence is critical for effective pruning and interpretability.

\section{Experiments}
\label{sec.experiments}
We conduct a series of experiments to investigate whether \texttt{ReGraMa} can mitigate neuronal activity loss and enhance performance. Specifically, we evaluate the effectiveness of \texttt{ReGraMa} across three representative and widely adopted architecture types: (i) the residual network-based policy (\autoref{sec.6.1}), (ii) the online policy parameterized by a diffusion model (\autoref{sec.6.2}), and (iii) the MLP policy featuring various activation functions (\autoref{sec.6.3}). Finally, we verify the robustness of \texttt{ReGraMa} with respect to the threshold $\tau$ (\autoref{app.tau}).

\begin{figure}[h] 
\centering
\includegraphics[width=1\textwidth]{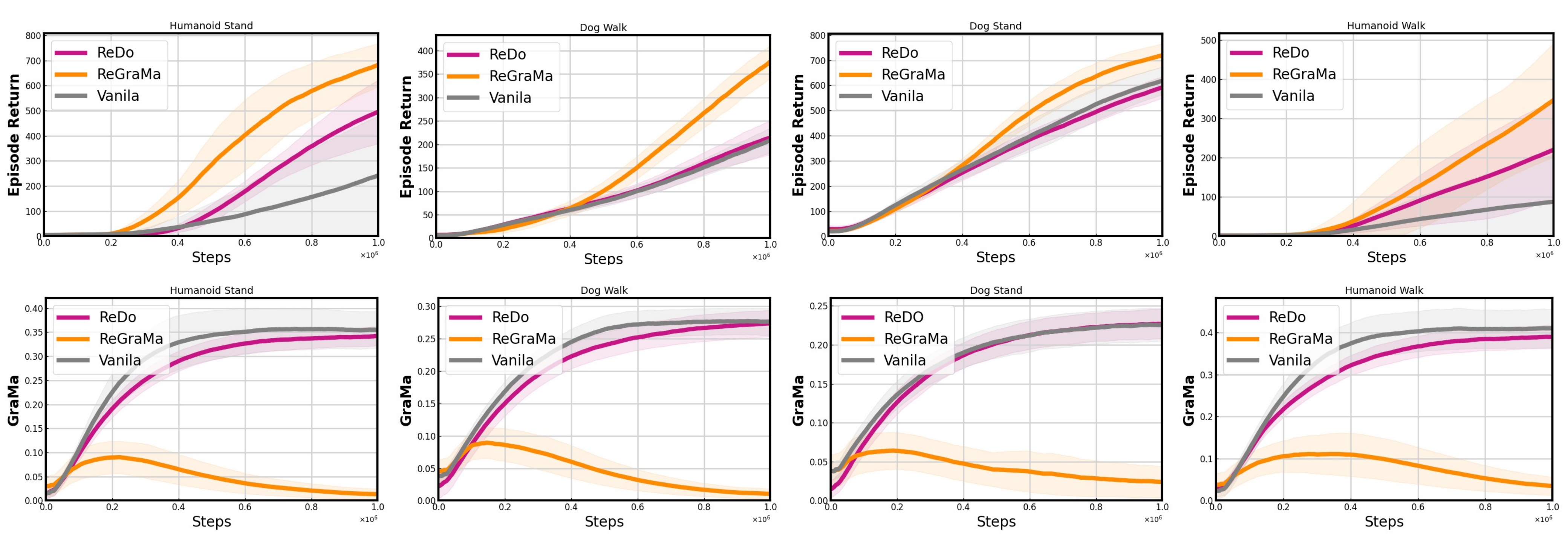}
\caption{\textbf{Performance and neuron inactivity with the default \texttt{BRO-net} size.}
(Top row) Episode return across four environments (\texttt{Humanoid Stand}, \texttt{Dog Walk}, \texttt{Dog Stand}, \texttt{Humanoid Walk}). (Bottom row) Corresponding proportion of inactive neurons. \texttt{ReGraMa} consistently achieves higher returns while maintaining fewer inactive neurons compared to \texttt{ReDo} and the vanilla baseline, demonstrating its effectiveness in stabilizing learning dynamics. Results are averaged over four seeds.}
\label{fig.8}
\end{figure}

\subsection{Residual Net-based Policy}\label{sec.6.1}
\paragraph{Experiment setup.}

Recent studies \citep{nauman2024bigger,lee2025simba} 
have shown that integrating residual modules into deep RL agents can significantly improve representation capability on complex visual tasks. 
We choose BRO-net \citep{nauman2024bigger} as a representative baseline and evaluate all methods on four challenging tasks from the DeepMind Control Suite \citep{Tassa2018DeepMindCS}. All the algorithm parameters follow the default settings. We set the empirical threshold $\tau=0.01$ for \texttt{ReGraMa}, and use the same ReDo's hyperparameters. The reset period is fixed at $1000$ steps. Further details are provided in Appendix \ref{app:exp_p_r}.
\paragraph{Main results.} Results in \autoref{fig.8} show that \texttt{ReGraMa} can accurately reset the neurons with low activity in each stream of the multi-branch network, thus effectively maintaining the learning capacity of the deep RL agent on the four complex tasks and enabling continual learning. In contrast, ReDo performs poorly. This performance gap arises because, as discussed in \autoref{sec.3.1}, ReDo misidentifies inactive neurons in multi-branch networks, undermining the effectiveness of its reset schedule.
\begin{wrapfigure}{r}{0.44\textwidth}
\centering
\vspace{-0.2cm}
\includegraphics[width=\linewidth]{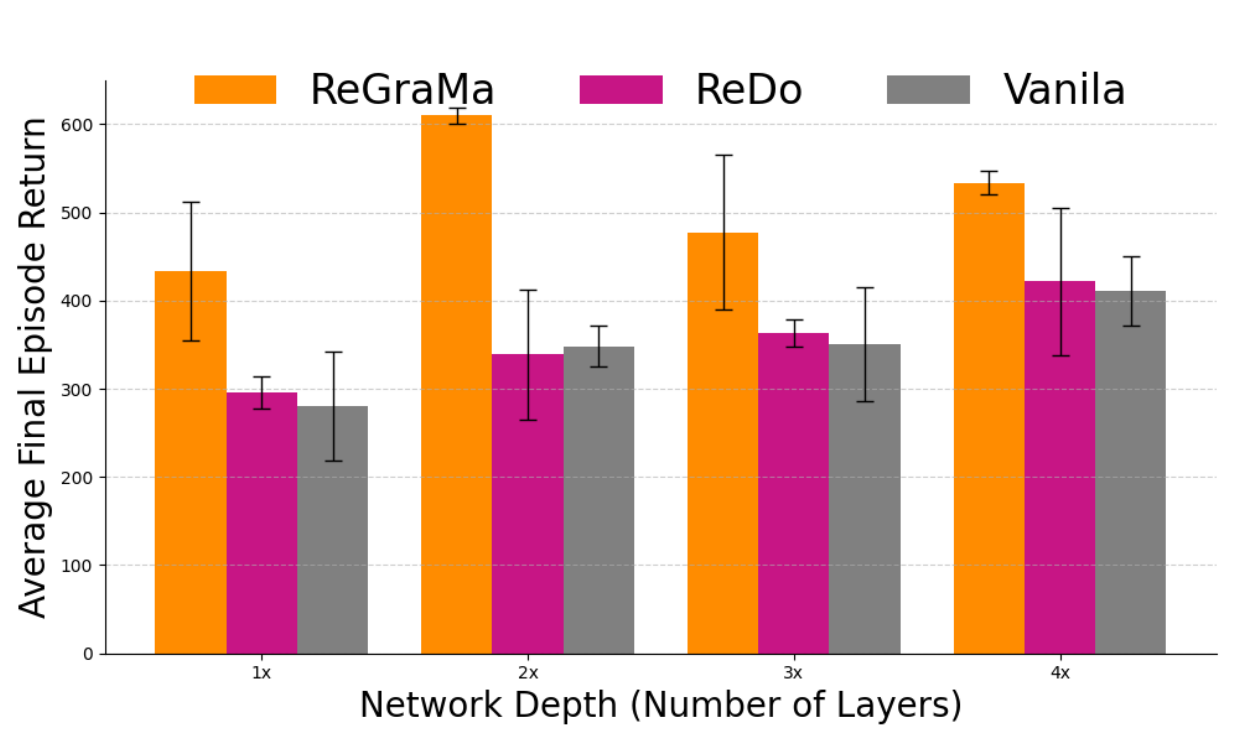}
\vspace{-0.7cm}
\caption{\textbf{\texttt{ReGraMa} is more robust under network scaling.} Results averaged over four DMC tasks with 12 seeds per method.}
  \label{fig.scale}
  \vspace{-1cm}
\end{wrapfigure}
\paragraph{Network scaling.}
To assess scalability, we evaluate both ReDo and \texttt{ReGraMa} under increased network depth using the \texttt{BRO-net} architecture across four challenging environments: Dog Walk, Dog Stand, Humanoid Walk, and Humanoid Stand. As the network grows deeper, training stability and gradient signal propagation typically become harder to maintain, often leading to degraded performance or inefficient utilization of added capacity. In this context, we analyze whether the gradient-based reset criterion of \texttt{ReGraMa} remains effective under larger model scales. 

Results show that \texttt{ReGraMa} consistently preserves stable performance improvements as network depth increases, demonstrating that its gradient-driven measure generalizes well across scales. In contrast, activation-based resetting (ReDo) fails to exploit the additional representational capacity, showing marginal gains at best and even degrading performance in the 2$\times$ model. These findings indicate that \texttt{ReGraMa} scales more gracefully with model size, effectively leveraging deeper architectures without introducing instability.

\subsection{Diffusion Model-Based Policy}\label{sec.6.2}
\paragraph{Experiment setup.}
\begin{wrapfigure}{r}{0.44\textwidth}
\centering
\vspace{-1.2cm}
\includegraphics[width=1\linewidth]{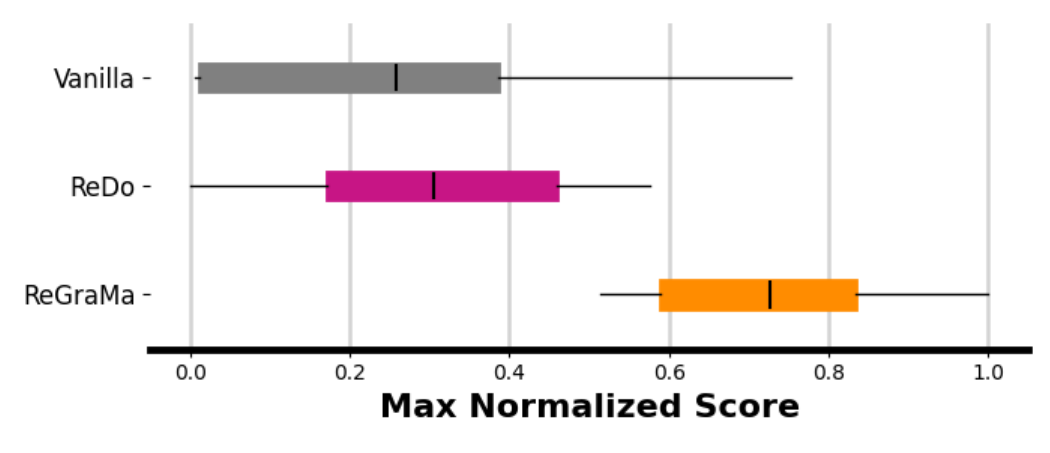}
\vspace{-0.7cm}
\caption{\textbf{Normalized scores for Ant and Walker2d.} Boxes show 4 seeds, whiskers indicate min/max, the midline denotes the median.}
  \label{fig.dacer_score}
  \vspace{-0.1cm}
\end{wrapfigure} Recent works have demonstrated that diffusion models, due to their strong expressiveness over multi-modal distributions, can significantly improve RL performance on complex control tasks \citep{Chi2023DiffusionPV}.
We use DACER \citep{wang2024diffusion}, a recent online diffusion policy, as a baseline and test ReDo and \texttt{ReGraMa} on two MuJoCo-v4 tasks from the original paper. DACER relies on a Unet backbone with Swish activations. All parameters follow official defaults. Hyperparameters for \texttt{ReGraMa} and ReDo are the same as in \autoref{sec.6.1}, and the reset period is fixed at 1,000 steps. Full details are in Appendix \ref{app:exp_p_d}.\begin{wrapfigure}{r}{0.55\textwidth}
\centering
\vspace{-0.8cm}
\includegraphics[width=\linewidth]{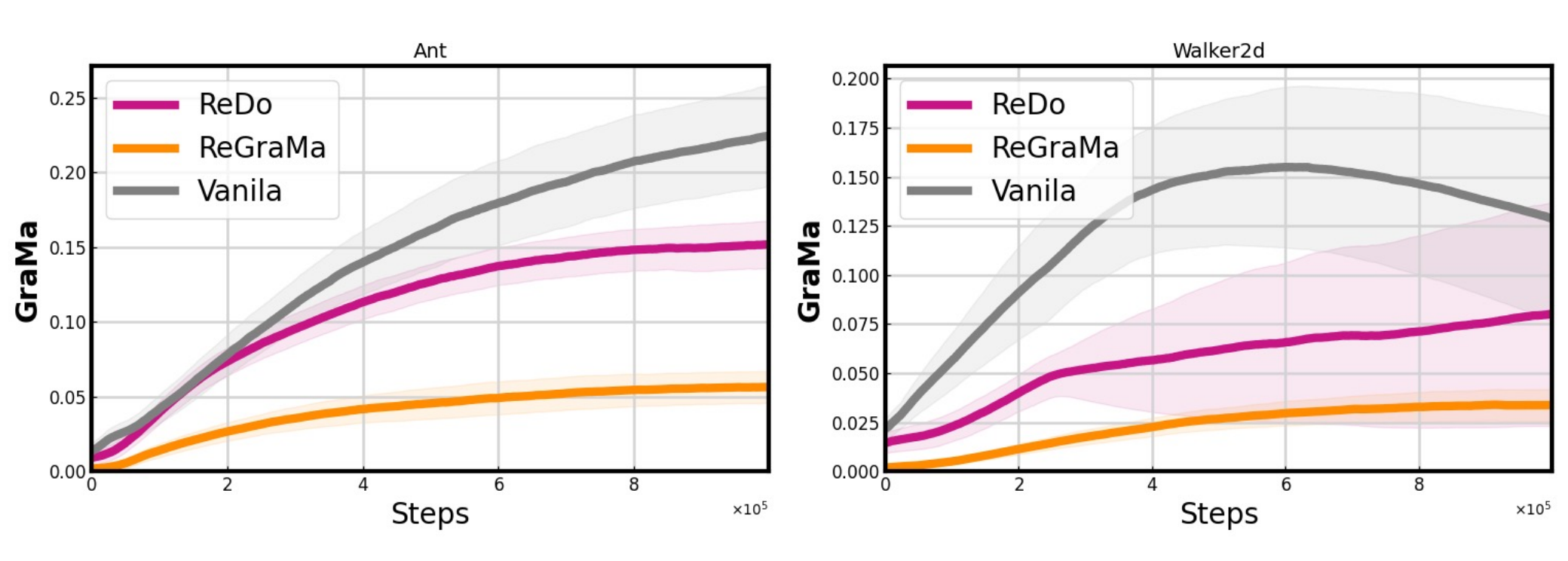}
\vspace{-0.75cm}
\caption{\textbf{Proportion of inactive neurons during training across two MuJoCo tasks} (\texttt{Ant} and \texttt{Walker2d}). \texttt{ReGraMa} maintains a consistently lower ratio of inactive neurons.}
  \label{fig.dacer_gate}
  \vspace{-0.5cm}
\end{wrapfigure}
\paragraph{Results.}As shown in \autoref{fig.dacer_score}, \texttt{ReGraMa} maintains robust and consistent performance on this complex architecture, whereas ReDo provides only marginal improvements. This discrepancy is explained by the ratio of inactive neurons in \autoref{fig.dacer_gate}, which reveals that ReDo fails to reset neurons that have lost learning capacity. In contrast, \texttt{ReGraMa} accurately identifies and resets low-activity neurons, preserving the agent's ability to learn and improving overall performance.

\subsection{Activation Function Variants}\label{sec.6.3}
\paragraph{Experiment setup.} 
\cite{nadimpalli2025evolving} shows that saturated activation functions in the hidden layers may improve RL agents' performance. Following their setup, we replace ReLU in SAC with Tanh and Sigmoid, keeping all other parameters as default. This enables us to evaluate the robustness of \texttt{ReGraMa} across various activation functions while minimizing the influence of extraneous factors on the experimental outcomes. We set $\tau=0$ for \texttt{ReGraMa}, and configure ReDo as in \autoref{sec.6.1}. All methods are evaluated on the challenging Ant task. Hyperparameter are provided in Appendix \ref{app:exp_p_sac}.

\begin{figure}[t] 
\centering
\vspace{-0.5cm}
\includegraphics[width=.9\textwidth]{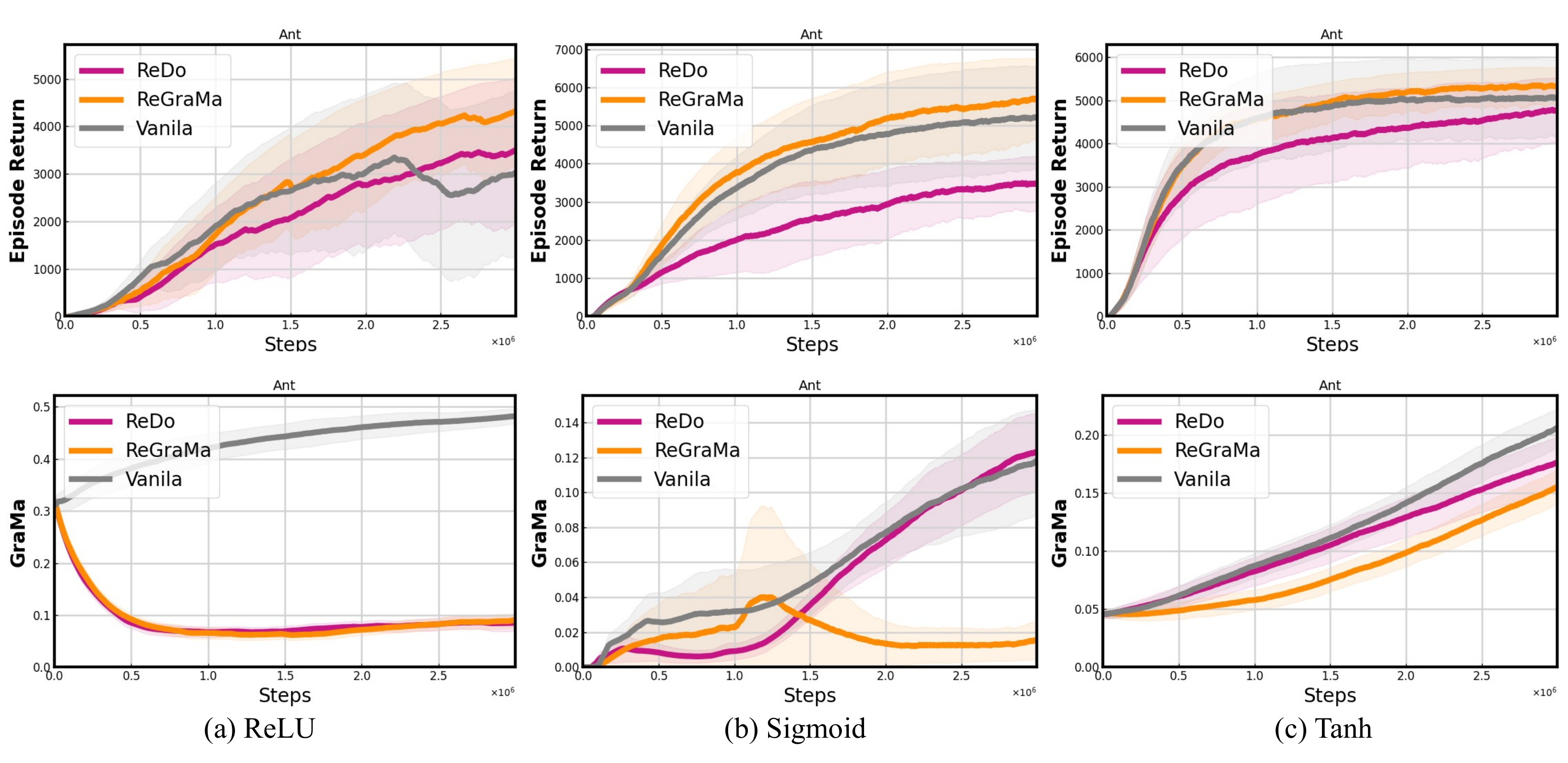}
\caption{\textbf{Effect of activation functions on performance and neuron inactivity.} (Top row) Episode return across training under different activation functions (\texttt{ReLU}, \texttt{Sigmoid}, \texttt{Tanh}). (Bottom row) Corresponding proportion of inactive neurons. \texttt{ReGraMa} maintains higher performance and lower inactivity levels across all activations, indicating that gradient-based resetting generalizes effectively beyond specific nonlinearities. Curves are averaged over four seeds.}
\label{fig.10}
\vspace{-0.2cm}
\end{figure}

\paragraph{Results.}
Results in \autoref{fig.10} show that \texttt{ReGraMa} accurately identifies low-quality neurons, mitigating neuronal inactivity and avoiding instability caused by false reset. While ReDo performs well under the ReLU, its performance degrades significantly with other activation functions, sometimes failing below vanilla SAC. In \autoref{fig.10}(c), although \texttt{ReGraMa} outperforms ReDo under the Tanh activation function, the proportion of inactive neurons gradually increases during training. We leave further investigation into whether this behavior stems from the perspective of the activation function or reset mechanism to future work.
\section{Discussion}
\label{sec:conclusion}

This research focuses on the critical issue of neuronal activity loss during training in deep RL agents. We show that the commonly used $\tau$-dormant neuron ratio (ReDo), which relies on neuron activations, struggles to capture learning activity in modern, highly parameterized agents. This limitation arises because activation sparsity does not directly reflect a neuron's contribution to learning. We shift focus from activations to gradients, and introduce \texttt{GraMa}, a simple, efficient, and architecture-agnostic metric based on gradient magnitude. \texttt{GraMa} enables accurate tracking of neuron-level learning dynamics across a broad range of network types, including residual and diffusion-based policies, and substantially recovers learning activity via guiding neuron resetting (\autoref{sec.3.2}). Our findings suggest that even high-capacity policies in deep RL suffer from underutilization at the neuron level, which may hinder generalization, multi-task transfer, and continual adaptation as in supervised learning \citep{Dohare2024LossOP}. By providing a lightweight diagnostic and intervention tool, \texttt{GraMa} opens the door to more principled approaches for maintaining continuous learning ability in deep RL agents.

\paragraph{Limitations.} Our evaluation is limited to three representative neural architectures due to computational constraints. Future work will extend \texttt{GraMa} to more complex settings, including large-scale transformer policies, multi-task \citep{mclean2025mu,nauman2025bigger}, and continual deep RL \citep{tang2025mitigating}. We aim for \texttt{GraMa} to serve as a practical tool for diagnosing and preserving learning capacity in deep RL agents.

\paragraph{Acknowledgment}
This work is supported by the National Natural Science Foundation of China 62406266. The authors would like to thank the reviewers for providing valuable feedback on the paper. We would also like to thank the Python community \cite{van1995python, 4160250} for developing tools that enabled this work, including NumPy \cite{harris2020array}, Matplotlib \cite{hunter2007matplotlib}, Jupyter \cite{2016ppap}, and Pandas \cite{McKinney2013Python}.

\paragraph{Broader Impact} This paper presents work whose goal is to advance the field of Machine Learning. There are many potential societal consequences of our work, none which we feel must be specifically highlighted here.

\bibliographystyle{plainnat}
\bibliography{references}

\clearpage
\appendix
\section{Experimental Details}\label{app:exp}
\subsection{Residual network based policy}\label{app:exp_p_r}
BRO-net  \citep{nauman2024bigger} is the first model-free algorithm to achieve near-optimal policies in the notoriously challenging Dog and Humanoid tasks. Its residual module-based architecture also gives it a powerful ability to scale up, making it widely concerned \citep{Lee2025HypersphericalNF, lee2025simba}. To this end, we choose BRO-net as the advanced agent for the multi-branch architecture to test the effectiveness of our metrics. Our implementation of BRO-net is based on the official implementation  \footnote{\url{https://github.com/naumix/BiggerRegularizedOtimistic_Torch}}. 

\paragraph{BRO-net Architecture}
The detailed structure of the core block used in BRO-net is shown in \autoref{tab:bro_net_block}.  The comprehensive Actor-Critic structure build by the above blocks is outlined in \autoref{tab:brotorch_actor_critic}.

\begin{table}[h!]
\centering
\caption{BroNetBlock Structure. $H_b$ denotes the block's internal hidden dimension (e.g., 256).}
\label{tab:bro_net_block}
\begin{tabular}{ll}
   \toprule
   Step & Layer Configuration \\
   \midrule
    1. FC Layer       & Linear($H_b, H_b$) \\
    2. Norm + Act     & LayerNorm($H_b$), ReLU \\
    3. FC Layer       & Linear($H_b, H_b$) \\
    4. Norm           & LayerNorm($H_b$) \\
    5. Residual       & Output = Step 4 Output + Block Input \\
   \bottomrule
\end{tabular}
\end{table}

\begin{table}[ht]
\centering
\caption{Whole network architectures. The structure of BroNetBlock is detailed in Table~\ref{tab:bro_net_block}.}
\label{tab:brotorch_actor_critic}
\begin{tabular}{lll}
   \toprule
   Layer & Actor Network & Critic Network (per Critic) \\
   \midrule
    Fully Connected & (state dim,256) & (state dim + action dim, 256)  \\

    LayerNorm & LayerNorm  & LayerNorm\\

    Activation & ReLU & ReLU\\

    BroNetBlock  & $N \times$ BroNetBlock  & $N \times$ BroNetBlock\\

    Fully Connected   &(256, 2 $\times$ action dim) & (256, 1) \\

     Activation     & Tanh & None\\
   \bottomrule
\end{tabular}
\end{table}

\paragraph{Hyperparameter setting}
To ensure a fair and reliable comparison across methods, we follow a fixed and consistent hyperparameter selection protocol, as inconsistent hyperparameter choices are known to significantly affect conclusions in value-based deep RL \citep{ceron2024on}. The shared hyperparameters of the BRO-net algorithm utilized in all our experiments are outlined in \autoref{app:brotorch_hyperparams}. Both Redo and \texttt{Grama} were configured with the recommended values for $\tau$ ($0.01$ for \texttt{Grama} and $0.02$ for ReDo) and the same reset frequency (every $1000$ steps). To reproduce the learning curves shown in the main text, we advise using seeds ranging from $0$ to $4$. For the scale experiments, we increased the number of BroNetBlock from $1$ to $4$ in both actor and critic. 
\begin{table*}[ht]
\caption{Hyperparameter settings for BRO}
\centering
\begin{tabular}{ll} 
   \toprule
   Hyperparameter & Value  \\
   \midrule
    Actor Learning Rate & $1 \times 10^{-4}$ \\
    Critic Learning Rate & $1 \times 10^{-3}$  \\
    Replay Ratio & 2  \\
    Discount Factor ($\gamma$) & $0.99$  \\ 
    Batch Size & $128$ \\
    Buffer Size & \num{1e6} \\
    Actor BroNetBlock & $1$ \\
    Critic BroNetBlock & $2$ \\ 
    \midrule
\multicolumn{2}{c}{\textit{Reset Specific Parameters}}\\
    \midrule
    Reset $\tau$ & $0.01$ \\
    Reset Frequency & $1000$ \\
   \bottomrule
\end{tabular}
\label{app:brotorch_hyperparams}
\end{table*}
\subsection{Diffusion model based policy}\label{app:exp_p_d}

\paragraph{Network Architecture}
As one of the recent works to successfully construct online policies based on the diffusion model, Dacer has received extensive attention from the community and has been used as a baseline by some recent studies \citep{Dong2025MaximumER, Ma2025SoftDA}. To test the effectiveness of our metrics in the advanced diffusion model policies, we chose the official Swish activation function and U-net based Dacer as a baseline with complex architecure. We reproduce the version of the code that introduces the two neuronal metrics into the policy model, based on the official code of DACER\footnote{https://github.com/happy-yan/DACER-Diffusion-with-Online-RL}.
Our detailed Structures were showen in \autoref{app:dacer_structures_symbolic2}.

\begin{table}[ht]
\centering
\caption{Network Structures for DACER.}
\label{app:dacer_structures_symbolic2}
\begin{tabular}{lll}
   \toprule
   Layer &  Actor Network &  Critic Network  \\
   \midrule
    Fully Connected & (state dim + time embedding dim, 256) & (state dim + action dim, 256) \\
    Activation & ReLU & ReLU \\
    Fully Connected & (256, 256) & (256, 256)\\ 
    Activation & ReLU & ReLU \\
    Fully Connected & (256, action dim) & (256, 2)\\
   \bottomrule
\end{tabular}
\end{table}

\paragraph{Hyperparameter setting} Our experiments adhere to the hyperparameter listed in \autoref{tab:dacer_hyperparameters}. $\tau$ for ReDo: $0.02$; $\tau$ for \texttt{Grama}: $0.01$; reset frequency (every $1000$ steps). To reproduce the learning curves shown in the main text, we advise using seeds ranging from $0$ to $4$.

\begin{table}[ht]
\centering
\caption{Hyperparameters for DACER Training}
\label{tab:dacer_hyperparameters}
\begin{tabular}{ll}
   \toprule
   Hyperparameter & Value \\
   \midrule
   Actor Learning Rate & $3 \times 10^{-4}$ \\
   Critic Learning Rate & $3 \times 10^{-4}$ \\
   Alpha Learning Rate & $3 \times 10^{-2}$ \\ 
   Discount Factor ($\gamma$) & 0.99 \\
   Batch Size & 256 \\
   Replay Buffer Size & $1 \times 10^6$ \\
   Target Network Update Rate ($\tau$) & $5 \times 10^{-3}$ \\
   Policy Update Delay & 2 \\
   Hidden Layer Size & 256 \\
   Reward Scale & 1.0 \\
   \midrule
  \multicolumn{2}{c}{\textit{Reset Specific Parameters}}\\
    \midrule
    Reset $\tau$ & $0.01$ \\
    Reset Frequency & $1000$ \\
   \bottomrule
\end{tabular}
\end{table}

\subsection{MLP-based SAC}\label{app:exp_p_sac}
\paragraph{Network Architecture} We use CleanRL for SAC (also Resnet SAC) implementation, which can be found at \url{https://github.com/vwxyzjn/cleanrl}. This library is a reliable open-source resource for deep reinforcement learning, designed in a PyTorch-friendly manner. And the detailed structure is shown in \autoref{app:dacer_structures_symbolic1}.

\begin{table}[ht]
\centering
\caption{Network Structures for SAC}
\label{app:dacer_structures_symbolic1}
\begin{tabular}{lll}
   \toprule
   Layer &  Actor Network &  Critic Network  \\
   \midrule
    Fully Connected & (state dim, 256) & (state dim + action dim, 256) \\
    Activation & ReLU & ReLU \\
    Fully Connected & (256, 256) & (256, 256)\\ 
    Activation & ReLU & ReLU \\
    Fully Connected & (256, 2× action dim) & (256, 1)\\
    Activation & Tanh & None \\
   \bottomrule
\end{tabular}
\end{table}

\paragraph{Hyperparameter setting} To ensure a fair and reliable comparison across methods, we follow a fixed and consistent hyperparameter selection protocol, as inconsistent hyperparameter choices are known to significantly affect conclusions in value-based deep reinforcement learning \citep{ceron2024on}. The shared hyperparameters for the SAC algorithm are summarized in \autoref{tab:sac_hyperparameters}. Note: We impose a maximum reset percentage limitation of 5\% exclusively for the Humanoid task.

\begin{table}[ht]
\centering
\caption{Hyperparameters for the SAC}
\label{tab:sac_hyperparameters} 
\begin{tabular}{ll}
\toprule
Hyperparameter & Value \\
\midrule
Total Timesteps & $3 \times 10^6$ \\
Replay Buffer Size & $1 \times 10^6$ \\
Discount Factor ($\gamma$) & 0.99 \\
Target Smoothing Coefficient ($\tau$) & 0.005 \\
Batch Size & 256 \\
Learning Starts & $5 \times 10^3$ \\
Policy Learning Rate & $3 \times 10^{-4}$ \\
Q-Network Learning Rate & $1 \times 10^{-3}$ \\
Policy Update Frequency & 2 \\
Target Network Update Frequency & 1 \\
Automatic Entropy Tuning & True \\
\addlinespace 
\midrule
\multicolumn{2}{c}{\textit{Reset Specific Parameters}} \\ 
\midrule
Reset $\tau$ & $0$ \\
Reset Frequency & $1000$ \\
Max Reset percentage & 5\% (Humanoid) \\
\bottomrule
\end{tabular}
\end{table}

\newpage
\section{\texttt{ReGraMa} is more robust to the threshold $\tau$}\label{app.tau}
\begin{figure}
\centering

\includegraphics[width=\linewidth]{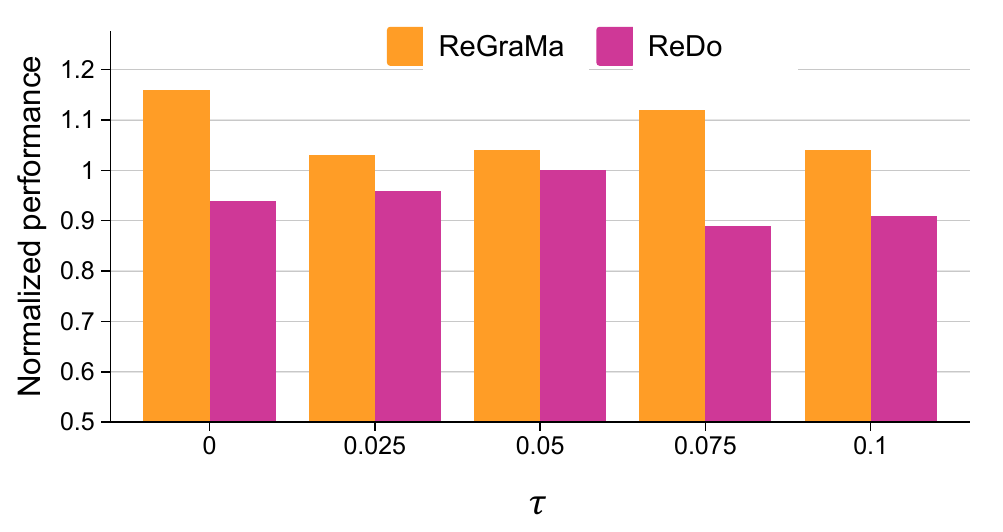}

\caption{Sample uniformly from $[0,0.1]$. Each bar is the average of 4 seeds.}
  \label{fig.tau}

\end{figure}
To assess the robustness of both metrics across varying thresholds, we use BRO-net as the backbone and evaluate performance on the challenging DeepMind Control Suite Humanoid Walk task. By systematically varying the $\tau$ following the recommended setup, we found that \texttt{ReGraMa} consistently outperformed ReDo (\autoref{fig.tau}). This proves that, even with relaxed restrictions, \texttt{ReGraMa} has less tendency to reset incorrectly.
\section{Additional experiments}
We further evaluate four challenging DeepMind Control (DMC) tasks with tensor-based observations. The results in \autoref{tab:ab_1} show that, under the BRO-net architecture, \texttt{ReGraMa} consistently outperforms all competing methods, highlighting its robustness in complex network architectures. ReBron \citep{qin2024the}, which explicitly accounts for both overactive and dormant neurons, also achieves strong performance, although it slightly degrades performance on the Dog Walk task. In contrast, S\&P \citep{Ash2019OnTD} and CBP \citep{Dohare2024LossOP} provide modest but consistent improvements across tasks.

We additionally include new ablation studies in the updated appendix to investigate the interaction between \texttt{ReGraMa} and commonly used regularization techniques. Following the optimal weight decay configuration reported in \citet{Lyle2024DisentanglingTC}, we evaluate the combination of \texttt{ReGraMa} with weight decay and L2 initialization ($\lambda = 10^{-2}$) on the DMC Quadruped Run task using DrQv2 \citep{Yarats2021MasteringVC} for 3M environment steps.

The results in \autoref{tab:quadruped_performance} indicate that combining \texttt{ReGraMa} with either L2 initialization or weight decay individually further improves learning efficiency. Among these variants, L2 initialization, which preserves plasticity by regularizing the initial parameters, appears more compatible with our method. In contrast, the simultaneous application of \texttt{ReGraMa}, L2 initialization, and weight decay does not yield additional gains. We leave a deeper investigation of the underlying interactions between \texttt{ReGraMa} and other regularization strategies to future work.

\begin{table}[htbp]
\centering
\caption{Normalized performance relative to the vanilla BRO-net policy on DMC tasks (mean $\pm$ std over 3 runs).}

\label{tab:ab_1}
\begin{tabular}{lcccc}
\toprule
Method & Humanoid stand & Humanoid Run & Dog stand & Dog walk \\
\midrule
vanilla BRO-net policy (baseline) & $1.0$ & $1.0$ & $1.0$ & $1.0$ \\
\textbf{ReGraMa} & $1.21 \pm 0.03$ & $1.16 \pm 0.08$ & $1.12 \pm 0.08$ & $1.08 \pm 0.04$ \\
ReDo & $1.17 \pm 0.07$ & $0.96 \pm 0.03$ & $0.92 \pm 0.12$ & $0.94 \pm 0.06$ \\
S\&P & $1.05 \pm 0.12$ & $1.08 \pm 0.04$ & $0.95 \pm 0.07$ & $1.07 \pm 0.02$ \\
ReBorn & $1.13 \pm 0.06$ & $1.05 \pm 0.07$ & $1.02 \pm 0.06$ & $0.91 \pm 0.13$ \\
CBP & $1.18 \pm 0.02$ & $0.99 \pm 0.12$ & $1.06 \pm 0.04$ & $1.09 \pm 0.07$ \\
\bottomrule
\end{tabular}
\end{table}

\begin{table}[htbp]
\centering
\caption{\textbf{Effect of combining \texttt{ReGraMa} with L2 initialization and weight decay} on DMC Quadruped Run (3M steps, DrQv2).}
\label{tab:quadruped_performance}
\begin{tabular}{lc}
\toprule
Method & Quadruped Run \\
\midrule
vanilla policy & $649.13 \pm 182.43$ \\
ReGraMa & $706 \pm 127.25$ \\
\textbf{\textit{ReGraMa + L2 init}} & $742.36 \pm 127.31$ \\
ReGraMa + weight decay & $751.31 \pm 94.26$ \\
ReGraMa + L2 init \& weight decay & $739.42 \pm 104.58$ \\
\bottomrule
\end{tabular}
\end{table}

\clearpage
\newpage
\section*{NeurIPS Paper Checklist}
\begin{enumerate}

\item {\bf Claims}
    \item[] Question: Do the main claims made in the abstract and introduction accurately reflect the paper's contributions and scope?
    \item[] Answer: \answerYes{} 
    \item[] Justification: Please refer to abstract and introduction.

\item {\bf Limitations}
    \item[] Question: Does the paper discuss the limitations of the work performed by the authors?
    \item[] Answer: \answerYes{} 
    \item[] Justification: The limitations can be seen in Conclusion of main paper.

\item {\bf Theory assumptions and proofs}
    \item[] Question: For each theoretical result, does the paper provide the full set of assumptions and a complete (and correct) proof?
    \item[] Answer: \answerYes{} 
    \item[] Justification: Please refer to Sec.4 of the main paper.

    \item {\bf Experimental result reproducibility}
    \item[] Question: Does the paper fully disclose all the information needed to reproduce the main experimental results of the paper to the extent that it affects the main claims and/or conclusions of the paper (regardless of whether the code and data are provided or not)?
    \item[] Answer: \answerYes{} 
    \item[] Justification: We make our code available.

\item {\bf Open access to data and code}
    \item[] Question: Does the paper provide open access to the data and code, with sufficient instructions to faithfully reproduce the main experimental results, as described in supplemental material?
    \item[] Answer: \answerYes{} 
    \item[] Justification: The code link can be found in abstrct, and the hyperparameter setting can be found in Appendix.

\item {\bf Experimental setting/details}
    \item[] Question: Does the paper specify all the training and test details (e.g., data splits, hyperparameters, how they were chosen, type of optimizer, etc.) necessary to understand the results?
    \item[] Answer: \answerYes{} 
    \item[] Justification: The experimental detail can be found in Experiment part. The hyperparameter setting can be found in Appendix B.

\item {\bf Experiment statistical significance}
    \item[] Question: Does the paper report error bars suitably and correctly defined or other appropriate information about the statistical significance of the experiments?
    \item[] Answer: \answerYes{} 
    \item[] Justification: The results are accompanied by error bars, confidence intervals. We detail the calculation of error bars, running steps and the number of seeds in the text description of each figure.

\item {\bf Experiments compute resources}
    \item[] Question: For each experiment, does the paper provide sufficient information on the computer resources (type of compute workers, memory, time of execution) needed to reproduce the experiments?
    \item[] Answer: \answerYes{} 
    \item[] Justification: The Experiments Compute Resources can be found in Appendix.
    
\item {\bf Code of ethics}
    \item[] Question: Does the research conducted in the paper conform, in every respect, with the NeurIPS Code of Ethics \url{https://neurips.cc/public/EthicsGuidelines}?
    \item[] Answer: \answerYes{} 
    \item[] Justification: We make sure to preserve anonymity.

\item {\bf Broader impacts}
    \item[] Question: Does the paper discuss both potential positive societal impacts and negative societal impacts of the work performed?
    \item[] Answer: \answerNA{} 
    \item[] Justification: This paper is only about promoting technological innovation and does not have social impact.
    
\item {\bf Safeguards}
    \item[] Question: Does the paper describe safeguards that have been put in place for responsible release of data or models that have a high risk for misuse (e.g., pretrained language models, image generators, or scraped datasets)?
    \item[] Answer: \answerNA{} 
    \item[] Justification: Ours paper poses no such risks.

\item {\bf Licenses for existing assets}
    \item[] Question: Are the creators or original owners of assets (e.g., code, data, models), used in the paper, properly credited and are the license and terms of use explicitly mentioned and properly respected?
    \item[] Answer: \answerYes{} 
    \item[] Justification: We cite the original paper that produced the code package, and state which version of the asset is used (include a URL). 

\item {\bf New assets}
    \item[] Question: Are new assets introduced in the paper well documented and is the documentation provided alongside the assets?
    \item[] Answer: \answerNA{} 
    \item[] Justification: This paper does not release new assets.

\item {\bf Crowdsourcing and research with human subjects}
    \item[] Question: For crowdsourcing experiments and research with human subjects, does the paper include the full text of instructions given to participants and screenshots, if applicable, as well as details about compensation (if any)? 
    \item[] Answer: \answerNA{} 
    \item[] Justification: Our paper does not involve crowdsourcing nor research with human subjects.

\item {\bf Institutional review board (IRB) approvals or equivalent for research with human subjects}
    \item[] Question: Does the paper describe potential risks incurred by study participants, whether such risks were disclosed to the subjects, and whether Institutional Review Board (IRB) approvals (or an equivalent approval/review based on the requirements of your country or institution) were obtained?
    \item[] Answer: \answerNA{} 
    \item[] Justification: Our paper does not involve crowdsourcing nor research with human subjects.

\item {\bf Declaration of LLM usage}
    \item[] Question: Does the paper describe the usage of LLMs if it is an important, original, or non-standard component of the core methods in this research? Note that if the LLM is used only for writing, editing, or formatting purposes and does not impact the core methodology, scientific rigorousness, or originality of the research, declaration is not required.
    \item[] Answer: \answerNA{} 
    \item[] Justification: NA

\end{enumerate}

\end{document}